# Deep Learning for Information Systems Research

Sagar Samtani, Assistant Professor and Grant Thornton Scholar, Kelley School of Business, Indiana University (ssamtani@iu.edu)

Hongyi Zhu, Assistant Professor, University of Texas at San Antonio, College of Business (hongyi.zhu@utsa.edu)

Balaji Padmanabhan, Professor, Director, Center for Creativity and Analytics, Muma College of Business, University of South Florida (bp@usf.edu)

Yidong Chai, Assistant Professor, Hefei University of Technology, School of Management, (yidongchai@gmail.com)

Hsinchun Chen, Regents' Professor, Director, Artificial Intelligence Lab, ACM, IEEE, AAAS Fellow, Eller College of Management, University of Arizona (hsinchun@arizona.edu)

# ABSTRACT

*Artificial Intelligence (AI) has rapidly emerged as a key disruptive technology in the 21$^{st}$ century. At the heart of modern AI lies Deep Learning (DL), an emerging class of algorithms that has enabled today's platforms and organizations to operate at unprecedented efficiency, effectiveness, and scale. Despite significant interest, IS contributions in DL have been limited, which we argue is in part due to issues with defining, positioning, and conducting DL research. Recognizing the tremendous opportunity here for the IS community, this work clarifies, streamlines, and presents approaches for IS scholars to make timely and high-impact contributions. Related to this broader goal, this paper makes five timely contributions. First, we systematically summarize the major components of DL in a novel Deep Learning for Information Systems Research (DL-ISR) schematic that illustrates how technical DL processes are driven by key factors from an application environment. Second, we present a novel Knowledge Contribution Framework (KCF) to help IS scholars position their DL contributions for maximum impact. Third, we provide ten guidelines to help IS scholars generate rigorous and relevant DL-ISR in a systematic, high-quality fashion. Fourth, we present a review of prevailing journal and conference venues to examine how IS scholars have leveraged DL for various research inquiries. Finally, we provide a unique perspective on how IS scholars can formulate DL-ISR inquiries by carefully considering the interplay of business function(s), application area(s), and the KCF. This perspective intentionally emphasizes inter-disciplinary, intra-disciplinary, and cross-IS tradition perspectives. Taken together, these contributions provide IS scholars a timely framework to advance the scale, scope, and impact of deep learning research.*

**Keywords:** *deep learning, artificial intelligence, information systems, knowledge contribution framework, information systems methodologies, research guidelines*

# Deep Learning for Information Systems Research

## INTRODUCTION

Artificial Intelligence (AI) has rapidly emerged as a key disruptive technology of the 21st century. Increasingly, many companies are capitalizing upon the initial promise of AI to conduct their business operations with unprecedented efficiency, effectiveness, and scale. The core engine that powers many AI innovations and development is Deep Learning (DL). In contrast to earlier AI approaches that relied on feature engineering or formal logic, DL can automatically process ever-increasing data sizes, continuously learn from rapidly evolving domains, and more closely emulate a human's thought processes and decision making. These capabilities have helped DL rapidly emerge as a viable solution for many high impact application areas. The widespread successes of DL have led to significant investments into developing novel research to solve grand societal challenges. Information Systems (IS) researchers have embraced DL, and more broadly, AI, as one of the key areas for growth in the discipline. This has been evident by the increased quantities of special issues, course developments, and community events related to the topic (Jain et al. 2018, Bardhan et al. 2020; Rai 2020a).

Despite significant attention on the topic, IS has not been viewed as a key driver or leader in DL-driven AI, and contributions of IS researchers in the DL area over the last ten years have been severely limited. This is attributable to several key challenges IS scholars face when aiming to make DL contributions. First, the delineation between how DL is instantiated within the context of IS versus related disciplines (e.g., computer science (CS)) has not yet been clearly defined. Consequently, how to formulate salient and high-impact DL research that does not directly compete with CS scholars remains a challenge. Second, IS scholars often face significant issues in positioning the contributions of their DL-based research. Oftentimes, IS scholars are faced with the challenge of balancing technical novelty and practical impact of their DL work. The compromises made in this regard can significantly hinder the viability of IS to make consistent contributions. Third, how to effectively conduct DL research for IS in a rigorous and repeatable fashion has not yet been formulated. As a result, it is very difficult for the IS discipline to quickly foster and grow researchers well-equipped in executing high impact DL research.

When taken together, these three challenges significantly impair IS researchers from being able to publish DL work and thereby severely limit the discipline from being viewed as a major player in DL research and practice. As a case in point, over the last five years (2016-2020) – arguably still the early phase of exponential grown for deep learning-driven AI – IS researchers have published 62 articles in IS journals and the International Conference on Information Systems (ICIS), as compared to over 86K articles in deep learning published by the larger computing community during this period as per the ACM Digital Library. A clear roadmap for how IS scholars can contribute in the DL area is necessary for the discipline to emerge as a leader within this major technological revolution. Unlike traditional machine learning (ML) that made developments over a span of 30-40 years, DL is progressing at a staggering rate on a day-to-day basis. Recognizing the tremendous opportunity here for the IS community, this work clarifies, streamlines, and presents approaches for IS scholars to make timely and high-impact contributions. To this end, this paper makes five timely contributions.

1. First, we formulate and present a novel Deep Learning for Information Systems Research (DL-ISR) schematic. In contrast to reviews commonly presented in major CS textbooks or in related course materials, this overview clarifies the role of DL within the context of IS. In particular, it illustrates how DL's technical procedures are driven by key business requirements from a particular application environment.

2. Second, we present a Knowledge Contribution Framework (KCF). This framework aims to provide IS scholars with a "big tent" mechanism to carefully articulate and position a wide array of DL-ISR contributions for maximum impact and visibility. As a result, it holds significant promise for allowing IS scholars to *rapidly* advance systematic streams of high-impact research.

3. Third, we carefully draw from principles within the scientific method, fundamental AI, and design science to articulate ten guidelines to assist IS scholars in executing rigorous and scientifically sound deep learning research. These guidelines were carefully constructed to help IS scholars showcase high-impact research while avoiding pitfalls in the review process.

4. Fourth, we provide a systematic review of the DL research published in prevailing IS journals and conferences. As part of this review, we provide a summary of the nature of the contributions via the KCF and compare the nature of the research inquiries examined against related disciplines.

5. Finally, we provide a unique perspective on how IS scholars can formulate DL-ISR inquiries. In particular, we illustrate how carefully considering the interplay of business function(s), application area(s), and the KCF can lead to DL research that naturally emphasizes inter-disciplinary, intra-disciplinary, and cross-IS tradition perspectives.

The remainder of this paper is organized as follows. First, we provide a systematic background of deep learning from the perspective of the IS discipline. Second, we summarize the proposed KCF, and detail its constituent components. Third, we describe the ten guidelines for presenting and positioning deep learning research in IS. Fourth, we summarize how the KCF can be leveraged to identify viable and promising deep learning research opportunities for IS scholars. The final section concludes this work.

## DEEP LEARNING BACKGROUND

DL is an area where both CS and IS scholars can make important contributions. However, these are likely to be different in nature. CS scholars primarily publish in conferences (faster review cycles, more opportunities), as opposed to journals (significantly slower and lengthier review cycles, fewer opportunities). Moreover, CS has historically focused on advancing the mathematical and theoretical foundations of DL. In contrast, IS has traditionally been an application-driven, information-centric discipline. Contributions are often situated implementations of a novel IT artifact that have been carefully designed and evaluated based on the unique data characteristics of a selected business or societal environment (Rai 2017).

Keeping these different perspectives in mind, we present a conceptual schematic of deep learning for information systems research (DL-ISR) in Figure 1. This schematic provides a holistic, visual, and abstracted view of how the DL engine is deeply integrated into an environment to provide AI-driven innovation. Four major components comprise the DL-ISR: (1) Application Environment, (2) Standard ML Processes, (3) Deep Learning Design[1], and (4) Business and Societal Outcomes. The core, technical deep learning component is color coded in gray.

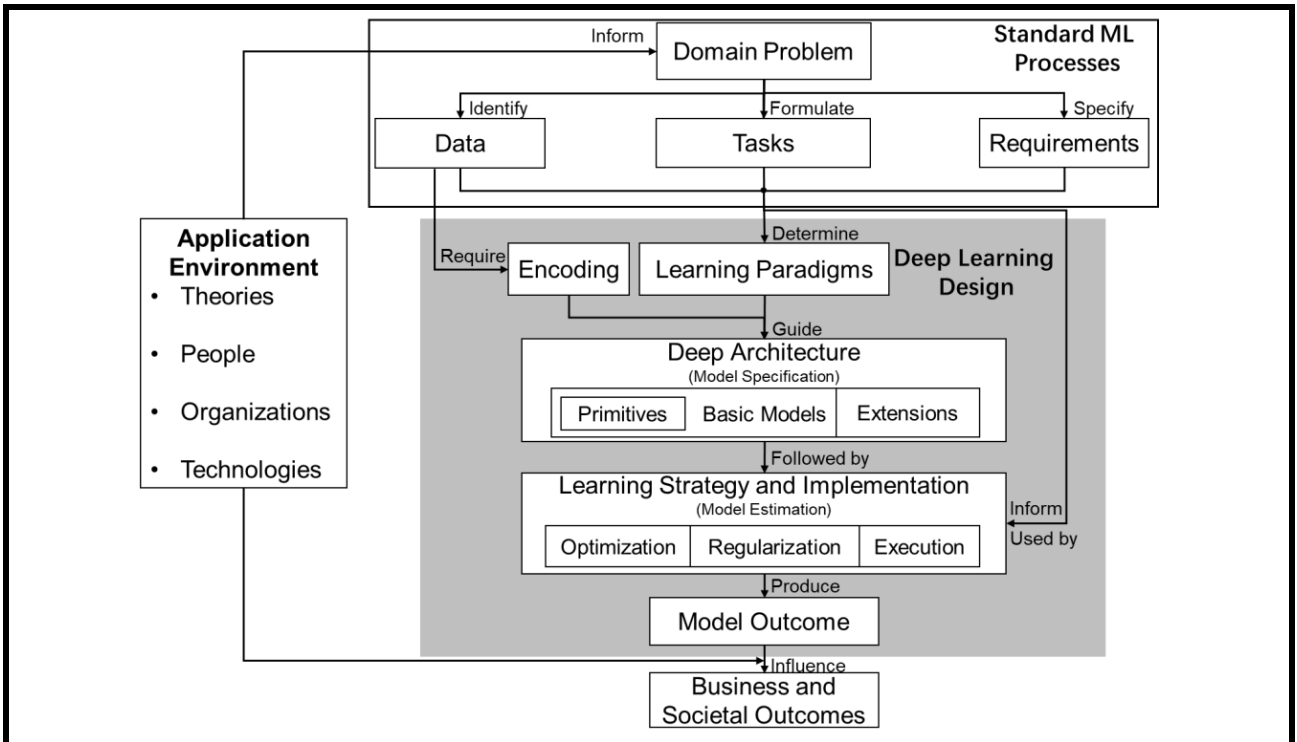

**Figure 1. Deep Learning in Information Systems Research (DL-ISR) Conceptual Schematic**

---

[1] Note that if the method of choice is conventional ML or other approaches (e.g., symbolic, Bayesian, econometric, etc.) instead of deep learning, then the deep learning box can be replaced. However, for reasons described in the introduction, we focus on deep learning in this paper.

Each major DL-ISR component has an associated set of concepts that are commonly considered by researchers. The application environment comprises theories, people, organizations, and technologies. Standard ML processes include data, tasks, and requirements. The DL component encompasses encoding, learning paradigms, architecture, learning strategy and implementation, and model outcome. Each helps drive business and societal outcomes. Concepts within and across components are linked through a series of relationships. The semantics of the relationships between concepts is summarized in Table 1. The bolded, italicized text represents the logical flow of activities researchers often follow in DL-ISR. To make the description more tangible, we provide a running example of how the component, concept, and relationship is instantiated within a recent major success made by the Google DeepMind research group published in Nature. This work developed a novel DL approach for early acute kidney injury (AKI) alerting by examining the knowledge embedded in electronic health records (EHR) (Tomasev et al. 2020).

| Table 1. Semantics of Relationships Between Major DL-ISR Components | | |
|---|---|---|
| **Component** | **Semantics of the Relationship Between Concepts** | **Running Example from Tomosev et al. 2020** |
| Application Environment | Application environment ***informs*** the problem/question | - The need to reduce clinician workloads and improve patient outcomes drives the problem of developing accurate and timely AKI alerts. |
| | Application environment ***influences*** outcomes | - While the DL engine provides AKI alerts, how this is embedded within a new treatment protocol will affect improvements in both patient outcomes and overall costs. |
| Standard ML Processes | Domain problem helps ***identify*** required data | - The problem requires a ground truth AKI dataset. |
| | Domain problem helps ***formulate*** tasks | - The tasks are formulated as predictive (AKI onset, laboratory testing results). |
| | Domain problem helps ***specify*** requirements | - Early predictions (24 hours prior) are critical. Some false positives are acceptable given the relative rarity (but known severity) of AKIs for patients. |
| Deep Learning Design | Data ***requires*** an encoding | - Structured EHR entries are encoded as sequences of events to capture temporal dependency. |
| | Data, Task, and Requirements ***determine*** learning paradigms | - The key requirements of obtaining kidney disease-relevant, more generalizable, and more robust representations while addressing under class imbalance issues drive the selection of multi-task learning. |
| | Encoding and Learning Paradigms ***guide*** the deep architecture (i.e., the model specification) | - Sequential event prediction leads to the selection of Recurrent Neural Network models.<br>- The embedding layer reduced input dimension to reduce training complexity and accelerate model learning and execution. |

| | Learning Strategy and Implementation (i.e., model estimation) **follows** Deep Architecture and **uses** and **is informed by** Data, Task, and Requirements | - Overall loss function consists of AKI prediction loss and seven other laboratory test predictions.<br>- $L_1$ regularization on the embedding layer is used to select the most salient features. |
|---|---|---|
| | Learning Strategy and Implementation **produces** the Model outcome | Model outcome is prediction of AKI onset within 24 hours. |
| Business and/or societal outcomes | Application Environment and Model Outcomes together **influence** business and/or societal outcomes | Model outcome is only a prediction. A process (determined by key decision makers in the hospital environment) around how the Google DeepMind AKI alert is used will eventually determine whether costs increase/decrease and if patient outcomes are improved. |

In the following subsections, we further discuss each component, concept, and relationship in detail. As we describe each, we expand on DeepMind's work as a running example. Whenever possible, we also provide examples of recent IS literature that illustrate selected concepts.

## Application Environment

Most application environments comprise theories, people, organizations, and technologies (Hevner et al. 2004). Moreover, it is generally well known in IS that the environment plays a significant role in influencing the questions posed as well as the outcomes of any information technology (IT) artifacts that firms develop. Each component of the environment, or a combination thereof, helps inform the fundamental domain questions or problems scholars seek to examine. Grounding research inquiry within key environmental issues keeps the problem space bounded (Nunamaker et al. 1990; Simon 1996). In the AKI detection paper, the problem is motivated by AKI's high prevalence in all inpatients in the US (1 in 5), the need to reduce the high death rates associated with related admissions, and the challenge of providing clinicians with intelligent, automatic, real-time, and personalized AKI monitoring.

## Standard Machine Learning (ML) Processes

Deep learning is fundamentally a branch of ML. As a result, ML principles serve as the foundation for successful deep learning projects. Within DL-ISR, standard ML processes begin with a key domain problem or question that is informed by the environment. This domain problem guides scholars in three key activities. First, the domain problem helps identify the relevant data within the environment. These data can be generated from multiple sources, including humans (e.g., text, sensory, sound, video, etc.), machines (e.g., smart grid, log files, etc.), or a combination of both (Chen et al. 2012). Second, the domain problem helps formulate the predictive and explanatory tasks. Predictive tasks train algorithms to execute a pre-specified goal for enhanced decision-making, automation, and prescriptive applications (Shmueli and Koppius 2011). Explanatory tasks aim to describe and explore data via unsupervised algorithms, summary statistics, visualizations, correlations, rules, and other mechanisms (Abbasi et al. 2016). Finally, the domain problem helps specify requirements essential for the environment. These can range from industry-specific regulations (e.g., HIPPA, GDPR), organization-specific processes (e.g., cost-effectiveness, privacy), end-user specifications (e.g., interpretability), and other constraints.

The emphasis on data identification, task formulation, and requirements specification can vary greatly based on their breadth, depth, evolution, and other environmental factors. As a result, this emphasis can help aide the articulation, evaluation, and ultimate instantiation of DL-based artifacts (Saar-Tsechansky 2015). Within the healthcare example, the data was carefully divided into broad segments to help increase the generalizability of the researchers' approach. Alternative approaches included grouping data into smaller units of analysis (e.g., elements within EHR). However, a higher granularity supported the requirements of early prediction on the scale of days. Other ML-based works within IS literature focused their research based on unique data characteristics (Zhu et al. Forthcoming), emerging regulatory requirements, and selected domain requirements.

### Deep Learning Design

DL belongs to an emerging branch of ML known as representation learning. DL methods can be fed raw data to automatically discover the representations of data required for selected learning tasks (e.g., supervised). A key reason for DL's rapid proliferation has been its ability to automatically extract multiple levels of representation (starting at the raw input) with a series of non-linear modules. This process requires considerably less human involvement than conventional ML feature engineering processes (LeCun et al. 2015). In general, DL design can be decomposed into five major components: encoding, learning paradigms, deep architecture, learning strategy, and model implementation and model outcome. Each is summarized and detailed in turn.

*Encoding*. Conventional ML methods require a carefully constructed set of feature vectors as input. While deep learning algorithms can operate on such representations, the algorithms' true power comes from automatically learning features from raw data. However, such learning still requires carefully pre-processing and encoding data into a structure recognizable by prevailing deep learning architectures. Encoding is the process of converting the raw data (a given set/sequence of characters, symbols etc.) into a specified format (Goodfellow et al. 2016). Each fundamental encoding format preserves particular types of information. For example, sequences can capture local and global temporal dependencies in text. Grids can encompass images, Internet of Things (IoT) device signals, temporal sequences (e.g., stock prices), and spatial data. Data sources can have multiple modalities, representations, and encodings simultaneously. Examples include video (spatial-temporal), attributed social networks (text, image, graph), and others. Encoding selection varies based on available data, key structural properties researchers wish to capture for their application, and/or relevant kernel theories. Within the AKI example, the encoding was developed by organizing EHR data sequentially based on time. This structure retained the real-world requirements for their predictive analytics task. Recent examples in IS literature have encoded text with rule-based features for text classification (Li et al. Forthcoming) and combined geographical and social data for personalized recommendations (Guo et al. 2018).

*Learning Paradigms.* There are numerous ML paradigms beyond the traditional predictive and explanatory dichotomy. For example, one could go with a traditional supervised learning paradigm or transfer learning paradigm (Long et al. 2015; Xu et al. 2017), which is built upon the concept that an existing model can be reused intelligently with new data. Within DL-ISR, the combination of data, tasks, and requirements determines how the learning paradigm is operationalized. For example, the data characteristics (e.g., richness of labels, volume) dictates the selection of semi/supervised, unsupervised, multi-view, multi-source, and online learning. Precise task specifications such as handling multiple datasets simultaneously, transferring knowledge across domain tasks, rewarding correct algorithmic behavior, and others, can help in selecting a finer-grained learning paradigm. These include multi-task (Collobert et al. 2008), transfer, reinforcement (Minh et al. 2015), and variational inference (Zhang et al. 2019). Finally, end-user requirements such as stability, privacy preservation, real-time processing, human involvement, interpretability, and others can motivate the selection of adversarial (Goodfellow et al. 2014), federated (Yang et al. 2019), active (Deodhar et al. 2017; Zheng & Padmanabhan 2006), and/or interpretable learning strategies. When considering the AKI context, the predictive (supervised) model operates in a multi-task fashion to capture and model multiple inputs simultaneously. The overall loss function consists of AKI prediction loss and seven laboratory-test predictions.

*Deep Architecture.* Encodings such as sequences, grids, and graphs require model designs (i.e., architectures) with the capacity to transform and manipulate such structures. Deep architectures generally comprise three building blocks: primitives, basic processing units, and extensions. Primitives are the core ingredients of deep learning and include neurons, activation functions, layers, weights, and error correction. Primitives can be organized into various functioning groups to form basic processing (i.e., neural) units. Each unit can operate on specific data encodings.

Prevailing units include the multi-layer perceptron (MLP) for conventional feature vectors, convolutional neural network (CNN) for grid-structured data, recurrent neural network (RNN) for sequential data, and graph neural network (GNN) for network structures. Extensions to these units increase their capacity to learn. Common extensions include capturing forward and backward processing for the bi-directional long short-term memory (BiLSTM), attention mechanism (Du et al. 2019; Rai 2020) integration for enhanced model interpretability, and others.

Architecture selection is guided by the learning paradigm and data encoding. Each basic processing unit can be extended to account for additional data characteristics, task specifications, and environment requirements. When considering the AKI example, the model was a novel long short-term memory (LSTM) customized to incorporate key EHR data characteristics. While the fundamental contributions in architecture primitives and basic processing units have originated from the CS community (LeCun et al. 2015), IS scholars have presented application-inspired extensions in recent work. Examples include extending word2vec's objective function (Shin et al. Forthcoming) and expanding a CNN's kernel function to capture interaction and sequential dependencies from sensor data (Zhu et al. Forthcoming).

*Learning Strategy and Implementation.* A strategy of how and what the model learns is specified after selecting the data encoding, learning paradigm, and deep architecture. Analogous to model estimation in conventional ML, the learning strategy is how the model is trained to execute the task. Strategies can focus on maximizing performance with novel loss functions, optimizing gradient descent through batch normalization, dropout, joint training, generalizing or regularizing the architecture with noise injections, parameter restrictions, and specifying activation functions and number of layers. Each can improve a model's performance and generalizability. The learning strategy and overall DL process is implemented into hardware and software based on the data, tasks, and requirements. Past IS scholars have defined custom loss functions to operate on multiple architectures simultaneously in a multi-task fashion (Ahmad et al. 2020), while others have discussed how their deep learning process can be implemented in real-time, distributed, and edge scenarios (Zhu et al. Forthcoming).

*Model Outcome.* Results from a deep learning procedure can oftentimes be the conclusion of an analytics-driven research project or serve as input for a subsequent process. The former often manifests itself in the form of predictive analytics, wherein a supervised learning model outputs a categorical or continuous value. The prediction of AKI is an example in this regard. However, deep learning outputs can also be a final or intermediate data representation (e.g., embedding, independent variable), pattern, synthetic data, or something else. Each can be integrated into subsequent (i.e., downstream) technical post-processing tasks, larger conventional ML workflows (e.g., RapidMiner or WEKA style), econometric models, and more. The model outcome is the final outcome at the end of a complete technical workflow. Figure 1 captures this by noting that the core DL engine components help "produce" the model outcome, and are not necessarily the model outcome themselves. However, they can serve as such, if there are no other workflows.

### Business and/or Societal Outcomes

Attaining strong model outcomes (e.g., performance, quality, etc.) does not necessarily lead to the optimal business outcomes (Gupta 2018). In contrast to CS, IS has a significant focus on developing technologies (e.g., algorithms, systems, etc.) that relevant stakeholders can adopt to increase productivity, reduce costs, save lives, and realize other societal benefits. Consequently, achieving maximal business and/or societal outcomes is contingent upon how the environmental factors (e.g., theory, processes, people, and technologies) and the model outcome interface. These factors can inform how a deep learning approach is designed and adopted within an environment.

From the design perspective, relevant computational theories or domain requirements can help guide the development, iterative refinement, and technical deployment of a deep learning process. When considering adoption, organizational processes, strategic policies, governance, and social science theories often help IS scholars identify how a novel deep learning-based algorithm or system is successfully adopted within a workplace. Similarly, relevant human computer interaction (HCI) theories combined with employee preferences serve as the foundation for how deep learning results can be presented in a user interface (UI) for maximal utility and benefit.

## KNOWLEDGE CONTRIBUTION FRAMEWORK (KCF) FOR DL-ISR

Each component and their constituent concepts of the DL-ISR schematic has substantial room for development. Advances in each component can be driven by many sources of inspiration within and across application environments. For example, the convolutional and pooling layers in the CNN was developed by emulating the processes of the human visual cortex (LeCun et al. 2015). Sources of inspirations can include data-generating processes, data characteristics, kernel theories, and principles in psychology, neuroscience, and biology. Similarly, learning paradigms have been designed to emulate human behavior. This includes adversarial learning, human negotiation, multi-agent environments, game theory, and others, can guide deep learning design. Unique data characteristics of an environment (e.g., labels, speed, veracity, etc.) can also be leveraged create novel advances. The wide range of possible motivations and sources of inspiration can lend itself to numerous DL innovations.

Despite significant potential, IS scholars currently lack a mechanism to clearly articulate the contributions of their deep learning research. Only 62 papers were published in prevailing IS journals and ICIS in 2016-2020 compared to 86K results for DL-related publications in the ACM Digital Library in the same time period. While the reasons for this are manyfold, an important issue is how reviewers assess DL contributions. Providing a framework for articulating contributions is paramount to accelerating the growth of an emerging research methodology or paradigm (Hevner et al. 2004). Therefore, we propose a novel Knowledge Contribution Framework (KCF) specifically for DL-ISR. This taxonomy groups major areas of the DL-ISR conceptual schematic in Table 2 to provide a lens for IS scholars to carefully position their deep learning research.

The KCF comprises five major contribution types: (1) Domain-specific, (2) Representation, (3) Learning, (4) System, Framework, or Workflow, and (5) Innovation-Accelerating. Contributions can solely reside in a single type or span multiple types. Table 2 summarizes each contribution type, definitions, and selected examples, and provides a mapping to the key elements in the conceptual DL-ISR schematic presented earlier in Figure 1.

| Table 2. Proposed Knowledge Contribution Framework (KCF) for DL-ISR | | | |
|---|---|---|---|
| **Contribution Type** | **Definition(s)** | **Mapping DL-ISR Conceptual Schematic** | **Sample References from IS Scholars** |
| Domain-specific | - Formulating and executing DL in a new application domain<br>- Studying how the environment influences business outcomes given a DL engine | - Environment<br>- Problem<br>- Data, task, and requirements | Ebrahimi et al. 2020; Li et al. 2016 |
| Representation | - Representing raw data in a novel structure for DL; or a new architecture of the underlying model for DL | - Encoding<br>- Deep architecture<br>- Basic processing unit<br>- Primitives<br>- Extensions | Ahmad et al. 2020; Zhu et al. 2020; Zhu et al. Forthcoming |
| Learning | - Developing new methods/techniques for learning/estimating DL models or developing novel specifications of the important components (e.g., a new loss function) | - Learning strategy<br>- Optimization<br>- Regularization<br>- Execution<br>- Learning paradigm | Ahmad et al. Forthcoming |
| System, Framework, or Workflow | - Creating a new way of integrating DL in a broader, higher-impact system<br>- Studying issues around the integration of DL systems | - Model outcome<br>- Business and societal outcomes | Zhang et al. 2020; Liu et al. 2020; Samtani et al. 2020 |
| Innovation-Accelerating | - Developing mechanisms to enhance scientific reproducibility of proposed deep learning approaches; sharing code and data | - Environment<br>- Business and/or societal outcomes | Zhu et al. Forthcoming; Samtani et al. 2020 |

Grouping contributions into these major categories can help authors, reviewers, and editors to quickly ascertain the novelty of deep learning-based research. As a result, the KCF holds significant potential in facilitating rapid advances in DL-ISR. In the following subsections, we further describe each contribution type, provide exemplar articles, and summarize common pitfalls. Guidelines on how to conduct DL-ISR research in a rigorous, relevant, and grounded fashion for each contribution type are presented in the following section.

*DL-ISR Contribution Type 1: Domain-Level Novelty*

IS scholars excel in asking innovative, application-driven research questions. Taking this into account, the first major type of DL-ISR contribution scholars can make occurs at the domain level. These contributions formulate and execute DL processes in a new application area. Two major categories of domain-level contributions can be made. The first pertains to applying emerging deep learning architectures (e.g., capsule) and/or learning paradigms (e.g., self-supervised) in a new, high-impact domain that has traditionally leveraged conventional ML or non-automated analyses. Such applications can yield significant performance gain and result in high-impact outputs (e.g., AKI research). Moreover, they can lead to valuable behavioral studies that study how factors within an environment (e.g., theories, processes, organizations, and other technologies) interact with DL-based model outcomes. The second type of contribution focuses on formulating new questions within a domain that were previously non-trivial without deep learning. Examples include formulating recommender systems that account for temporal sequences of customer preferences, identifying algorithmic gender or racial biases through interpretable deep learning, and others. Oftentimes, these contributions can pave the way for future scholars to execute research to attain deeper levels of novelty (e.g., representation, learning).

At first glance, achieving domain-level novelty would appear attainable. In reality, there are three key pitfalls in executing such research. First, contributions can be marginal, incremental, and/or obvious. This often occurs when the domain is well-studied or does not require DL. Second, incorrect DL processes can be selected. This often manifests itself when scholars acquire new datasets and directly apply well-established DL techniques without carefully considering the data characteristics, tasks, or domain requirements. Finally, contributions can often stay at the technical discussion (e.g., DL outperforms standard ML) without carefully considering the value of the proposed processes to the domain. Work exhibiting these issues are commonly rejected at top IS journals (Rai 2017). Each issue can potentially be mitigated by articulating the importance of the domain, clearly establishing the state of the art in previous literature, formulating new questions specifically enabled by DL, justifying why DL is necessary, demonstrating non-trivial technical improvements, and attaining non-obvious domain improvements of value to relevant stakeholders.

*DL-ISR Contribution Type 2: Representation-Level Novelty*

Deep learning's success is often contingent upon how raw data from an environment is encoded or structured. These encodings help motivate carefully constructed architectures that can fully capture the encoding's semantics and the environment's underlying data-generating processes. Research studies that design, develop, and evaluate novel encodings or architectures are representation-level contributions. Novel encodings can be constructed based on domain or engineering considerations, computational theory (e.g., information, statistical learning, graph, etc.), and/or relevant social, behavioral, and economic (SBE) perspectives (e.g., speech acts, lexical semantics, etc.). Each can provide new approaches on representing standard raw data types in a more holistic or targeted fashion. Recent examples within IS include incorporating sentiment scores (Xie et al. 2018), encoding demographic variables (Ahmad et al. 2020), and capturing interaction and temporal dependencies simultaneously (Zhu et al. 2020). Novel architectures are often motivated by designing structures that lead to more effective model computations (i.e., improves the model's capacity to learn). Examples of significant architectural contributions include incorporation of carefully constructed attention mechanisms to enhance model interpretability, design of novel processing units (e.g., highway networks), and enhancing models with additional processing capacity (e.g., bidirectional processing, filters, etc.).

A key challenge to research aspiring to attain representation-level novelty includes insufficient grounding, whether in key domain requirements, data characteristics, or theory. The novelty offered may also be trivial, incremental, or marginal. Examples include concatenating one or two features to embeddings or incorrect encoding selections (e.g., representing a graph's adjacency matrix into an image-like grid). These issues can be proactively mitigated by clearly delineating, justifying, and demonstrating the performance benefits between the proposed encoding and the original. The key issue pertaining to the architecture's novelty is claiming that common engineering considerations are new. Examples include stacking layers, specifying the number of neurons, and others. While tuning these parameters and hyperparameters is essential for sensitivity or ablation analyses, they are common to all deep learning processes. These issues can be mitigated by demonstrating how the proposed architecture operates on multiple datasets within or across domains.

*DL-ISR Contribution Type 3: Learning-Level Novelty*

A key reason for deep learning's rapid growth has been its ability to automatically learn (i.e., estimate and fit parameters) from encodings and architectures. The learning-level novelty in the KCF for DL-ISR reflects the contributions IS scholars can make to learn from increasingly complex encodings and architectures. Contributions made at this level can be to the *paradigm* (e.g., supervised, unsupervised, transfer, etc.) or to the *strategy* (e.g., joint training, backpropagation, weight sharing, etc.). With regards to the paradigm, novelties include converting architectures to operate in alternate settings (e.g., converting supervised approaches to unsupervised), extending or merging existing paradigms to operate on emerging architectures (e.g., creating new transfer learning mechanisms for spatial-temporal models), or defining new approaches altogether (e.g., self-supervised learning). Innovations on the learning strategy enhance how an architecture learns from encoded data. Examples of learning strategy novelties include approaches to training multiple model architectures simultaneously (i.e., joint training), new loss functions for specialized architectures (an area in which IS researchers can thrive by linking these deeply to business objectives), optimizing extended objective functions (e.g., min-max function in a generative adversarial network), and others. Each can help optimize, regularize, and effectively execute parameters within an architecture.

Common issues in this category are threefold. First, engineering procedures considered in all DL research can often be mistaken as novel contributions. Examples include adjusting batch sizes, learning rates, and other common engineering tasks. Second, variations can often lack proper justifications or rationale. This often occurs when a learning strategy is devised without careful consideration of key domain requirements. Finally, scholars can often loosely couple components (i.e., train them separately) to achieve a larger goal. This specifically occurs when merging conventional ML models with deep learning strategies. While being potentially system- or framework-level contributions (described next), such approaches can require joint training techniques to tightly couple multiple models.

*DL-ISR Contribution Type 4: System- or Framework-Level Novelty*

Systems development has been at the heart of the IS discipline for decades (Nunamaker et al. 1990). Oftentimes, various components carefully intertwined together can result in an end research product greater than the sum of its parts. The system- or framework-level novelty in the KCF for DL-ISR speaks directly to this long-standing tradition in IS. These novelties focus less on each individual component and more on composing multiple techniques into a larger workflow. A recent example within IS literature includes the TheoryOn System (Li et al. Forthcoming). TheoryOn leveraged rule-based approaches to retrieve hypotheses from literature, RNN-based techniques to extract variables, SVMs to extract theoretical relationships, and lexical similarity analysis to identify synonymous relationships. A user interface (UI) was built upon these analytics to enable users to retrieve contents of interest in an ad-hoc manner. Such systems can serve as IT artifacts for behavioral researchers to study their adoption within a workplace.

Two key issues can arise when striving to attain system or framework novelty. First, each component can lack appropriate justification or grounding. This can be mitigated by clearly articulating the strengths, weaknesses, and past examples of success. Second, these works can face significant challenges in evaluation. Carefully designing a workflow requires evaluating each individual component in its entirety. Despite these challenges, this level of contribution can enable an inclusive research agenda for all IS traditions. For example, inputting variables obtained from deep learning models into econometric models can help further identify causality within emerging phenomena. Deep learning-based systems can also serve as viable mechanisms to conduct targeted user evaluations to test for adoption, impact, and governance.

*DL-ISR Contribution Type 5: Innovation-Accelerating Contributions*

The long-term viability of an academic discipline is contingent upon the consistent delivery of novel research. Unlike other lines of scientific inquiry, DL requires significant resources and computational infrastructure. Correctly demonstrating each novelty requires evaluating how the proposed approach performs compared to well-established benchmarks. However, many researchers often face significant hurdles in re-implementing algorithms others have proposed. Even when the algorithm is reimplemented, the attained results are rarely at the same level presented in the original paper due to a lack of details about parameters and their tuning. Each hurdle can significantly limit the growth of a nascent field (Dennis et al. 2020).

The final type of DL-ISR contribution in our proposed KCF aims to help address these issues. Two types of contributions can be made at this level: public releases of content and replication studies. The former pertains to releasing code, datasets (with dictionaries, characteristics, and provenance), null results, and other contents essential to the core analytics process. Publicly accessible code and datasets can allow scholars to quickly build on past work and develop, test, and refine *newer* contributions. Knowledge about null results can provide an excellent resource for scholars to learn from past efforts, save a tremendous amount of time, and avoid fruitless paths (Nunamaker et al. 2017). Taking these together can significantly accelerate the rate of innovation.

Conducting replication research on the other hand can offer a significant contribution to the larger field, but may seem to slow down short-term progress. However, they are essential for the long-term benefit of an academic discipline for three reasons. First, if a replication does not hold, additional work will aim to identify why (Dennis et al. 2020). Second, replications that are successful can point to promising directions for future research. Finally, executing replications can also help quickly onboard new Ph.D., DBA, MS, and undergraduate students into the field. *AIS Transactions on Replication Research (TRR)* aims to serve as a vehicle to facilitate scientific replication for the IS discipline. Increasing the volume of scholars trained in fundamentally correct DL research can significantly accelerate the rate of innovations.

## GUIDELINES FOR PRESENTING AND POSITIONING DL-ISR

The KCF provides a novel lens for IS scholars to systematically contribute to DL-ISR. However, many DL-focused papers often face unique difficulties in the IS review cycles. Common issues include correctness of methodology, whether the work is truly novel or "just an application," clarity on contribution, presentation of ideas, ensuring reproducibility, and generalizability of the approach. These issues underscore the critical need for a set of guidelines that help IS researchers efficiently and effectively navigate these challenges and publish their research in a rigorous, timely, and scientific fashion. To this end, we propose ten guidelines to help IS researchers present and position their DL-ISR for maximum impact. The proposed set of guidelines aims to balance research, practice, and IS tradition, and to differentiate the IS discipline from CS. The guidelines find their original grounding in the scientific method, design science research, and/or AI/ML research. However, they have been articulated to account for the unique aspects of DL. Timely guidelines in this fashion have indeed contributed to the rapid development of methodological contributions to the field (Gregor and Hevner 2013; Hevner et al. 2004; Venkatesh et al. 2013; Peffers et al. 2007).

The ten guidelines can be broadly summarized into four major categories: Setup, Solution, Study, and Synthesis. Setup pertains to the motivation and scientific and practical basis for the proposed DL research. Solution comprises of the core technical DL development and evaluation. Study focuses on reflecting deeply on the Setup and Solution, and positioning the novelty of the proposed DL approach via the KCF. Finally, Synthesis focuses on how the proposed DL research can be translated into the business or societal context it was motivated by. We present each category and their respective guidelines in Table 3. While these guidelines reflect emerging norms in the DL area, they are presented in a prescriptive manner to help both authors and review teams present or evaluate new ideas consistently. We also summarize each guideline's scientific grounding.

| Table 3. Guidelines for Executing Deep Learning for Information Systems Research | | |
|---|---|---|
| **Category** | **Guideline** | **Scientific Grounding*** |
| Setup | Motivate and provide a clear, crisp, unambiguous problem specification. | DSR |
| | Review and summarize key data generating processes and past work to justify deep learning. | - |
| Solution | Systematically present how the proposed (technical) deep learning processes were formulated and developed. | DSR |
| | Rigorously evaluate the proposed deep learning approach with appropriate technical and non-technical experiments. | SM/DSR |
| | Provide procedural details of the deep learning approach to facilitate scientific reproducibility. | SM |
| | Articulate clearly the deep learning contribution via the KCF for DL-ISR. | SM, DSR, AI |
| Study | Reflect and summarize the key takeaways of the DL process. | AI |
| | Discuss any potentially unexpected or unintended consequences of the proposed DL model and its implementation. | AI |
| | State any potential ethical concerns in how the deep learning approach was developed and in its potential usage. | AI |
| Synthesis | Articulate how the proposed deep learning work can be translated into practice or industry. | - |

**\*Note:** DSR = Design Science Research; SM = Scientific Method.

The proposed guidelines can serve as a backbone for any DL-ISR paper and provide authors with a mechanism to rapidly develop and showcase their high-impact, novel research while avoiding common pitfalls. Similarly, these guidelines can help reviewers quickly evaluate and compare DL-based research. Over time, the guidelines will help create a body of work significantly different from other technical disciplines such as CS as well. In the following subsections, we further describe each guideline in detail. We organize the discussion of the guidelines based on the higher-level category they belong to.

### Setup

*Guideline 1: Motivate and provide a clear, crisp, unambiguous problem specification.* A DL-ISR project commences once a clear problem specification has been motivated and articulated. Two levels of details should be included. The first pertains to the higher-level overarching problem motivation of the key societal, business concern, and/or environment being studied (e.g., overall importance of cybersecurity, healthcare, etc.). These details are essential to understanding the overall domain being studied and its value. The second, lower level focuses on the specific problem being studied within the overarching context. Clearly articulating the specific problem being studied is essential for keeping the scope bounded. More importantly, it guides how the overall DL process is formulated (e.g., data, task, and requirements) and evaluated. A common issue is expressing the problem clearly, whether the higher-level overarching problem or the lower-level specific one. However, both are essential to convincingly conveying the scope, scale, importance, and bounds of the research study. These details also impact how the subsequent guidelines are followed.

*Guideline 2: Review and summarize key data generating processes and past work to justify deep learning.* Common questions that arise during academic reviews pertain to the data being studied, why DL is needed, and appropriateness and completeness of the features being used. Scholars can proactively address these concerns by reviewing the data-generating processes of the application and past research examining these data (Rai 2020). Key details for the data generating processes include summarizing how the data comes into existence, their provenance, data dictionaries, relationships between attributes, causal linkages, quality, assumptions, uncertainties, relevant theoretical relationships, and common pre-processing steps. Reviewing past work can further summarize how researchers have leveraged these data characteristics for the domain and problem under consideration. Key dimensions of past works that can be examined include motivation, representation, task, architecture, learning strategy, and key results. Kernel theories and domain-specific heuristics can also be summarized (Gregor and Hevner 2013). Well-executed reviews can ground, motivate, inspire, and justify relevant deep learning encodings, architectures, learning strategies, paradigms, and appropriate evaluations.

**Solution**

*Guideline 3: Systematically present how the proposed (technical) deep learning processes were formulated and developed.* Attaining novelty is essential for progressing scientific research. Within DL research, this entails summarizing the key design logic of the proposed processes. Key activities include exploring and justifying candidate encodings, architectures, learning strategies based on the data, task, and requirements. Once DL developments are made, researchers can present the conceptual novelty through several mechanisms. These include presenting key notation, mathematical formulations, pseudocode, end-to-end examples, and side-by-side diagrams that summarize the key differences between the proposed approach and the baseline. These descriptions should also be accompanied by a summary of the key technical and non-technical benefits of the proposed approach over prevailing benchmarks or practices.

*Guideline 4: Rigorously evaluate the proposed deep learning approach with appropriate technical and non-technical experiments.* While Guideline 3 emphasizes the importance of *conceptually* showing the novelty, Guideline 4 necessitates that researchers also demonstrate the validity of the posited novelty (i.e., does it solve the problem being posed). Two approaches can be taken: technical and non-technical. Technical evaluations compare the performance of the proposed deep learning process against prevailing ML and deep learning benchmarks. Well-established performance metrics and statistical tests measure performances. Five components comprise a thorough, convincing technical evaluation: dataset, model training and testing, model performance benchmarking, post-hoc (i.e., post-model) evaluation, and interpretation and insights (i.e., technical case study). The key aspects of each major component are presented in Table 4.

**Table 4. Summary of Major Components in a Technical Evaluation for DL-based IS Research**

| Component | Key Aspects | Description | Example(s)** |
|---|---|---|---|
| Dataset | Ground-truth dataset construction | Labelled dataset used for model training and testing representative of the phenomena of interest | Complete dataset fully labelled by experts |
| | Train | Portion of data that is used to train the algorithm(s) | Randomly selected 80% of the ground truth dataset |
| | Development (i.e., tuning) | Portion of data that is used to tune the algorithm(s) | Randomly selected 10% of the ground truth dataset |
| | Testing | Portion of data that is used to test and evaluate algorithm performance | Randomly selected 10% of the ground truth dataset |
| Model Training and Testing | Hyperparameter selection | Selecting values to control the learning process | Grid-search, pre-optimized, or trained model |
| | Training strategy | How the proposed model is trained and the model parameters learned | 10-fold cross validation, hold-out, pre-trained model, training strategy based on tests for overfitting and underfitting |
| Model Performance Benchmarking | Performance metric selection | Metrics to evaluate the performance | Accuracy, precision, recall, F1, NDCG, MAP, MRR, homogeneity, NMI |
| | Evaluation against non-DL models | Proposed DL model vs non-DL-based models | Naïve Bayes, SVM, Decision Tree |
| | Evaluation against DL models | Proposed DL model vs prevailing DL-based models | CNN, LSTM, GRU, RNN, ANN |
| Post-hoc (i.e., post-model) evaluation | Sensitivity or ablation analysis | Internal analysis of DL model to interpret how model components contribute to overall performance | # of layers, activation functions, varying model components, counterfactual analysis |
| | Convergence speed | How quickly the model converges | Speed, computational complexity |
| | Model stability | How stable the model is in training, comparison, etc. | Validation loss, thresholding, statistical significance |
| Interpretation and Insights (Technical case study) | Examples of outperformance | Identify 1-2 instances within the ground-truth dataset that were correctly identified by the proposed method, but missed by the best competing benchmark | - |
| | Apply proposed DL on unseen data | - | Applying a transfer learning framework to categorize all hacker exploits in forums |

**\*Note:** Post-hoc evaluations are similar to robustness checks in econometrics research. However, running these robustness checks require significantly higher investment and resources. Therefore, review teams should have a well-grounded and justifiable request for authors to run these.

**\*\*Note:** ANN = Artificial Neural Network; CNN = Convolutional Neural Network; GRU=Gated Recurrent Unit; LSTM = Long Short-Term Memory; MAP = Mean Average Precision; MRR = Mean Reciprocal Rank; NDCG = Normalized Discounted Cumulative Gain; NMI = Normalized Mutual Information; RNN = Recurrent Neural Network; SVM = Support Vector Machine

The scale, scope, breadth, and depth of technical evaluations will vary based on DL contribution type. Scholars that claim that their proposed DL approaches have representation or learning level novelties will often employ more extensive technical evaluations to fully demonstrate their novelties. Conversely, system- or application-level novelties may rely more on non-technical evaluations to evaluate the usefulness, usability, and value of a deep learning process. Non-technical evaluations can help determine whether the proposed technical work solves the higher-level problem being motivated from the environment. Possible evaluation approaches include surveys, randomized field experiments, embedded simulations, case studies, interviews, focus groups, and ethnographies. Subjects included in each evaluation type should ideally be drawn from the environment for which the approach is designed for. Careful selection of approach and subjects can convincingly illustrate the potential impact and adoption of the deep learning approach. For example, field studies can offer a powerful mechanism for examining critical issues related to DL, including AI trust, security, and privacy issues.

We do not advocate that both technical and non-technical approaches are *required* to demonstrate the validity, value, and usefulness of the proposed techniques. Non-technical evaluations may not be feasible or may significantly delay the publication of high-impact research. These situations are particularly common for healthcare and cybersecurity applications (Saar-Tsechansky 2015). Similarly, technical evaluations may not be required if the algorithms being deployed are well-accepted algorithms for specific contexts, or if the DL component is not the centerpiece. However, researchers need to have a clear rationale from past literature, best practices, and environmental requirements for their evaluation processes.

*Guideline 5: Provide procedural details of the deep learning approach to facilitate scientific reproducibility.* A key contribution that can be made from a DL study is the provision of their research materials. Ideally, this is executed by providing the code and data, with appropriate documentation to verify that the reported performances are true and correct. In cases where such provisions are not a possibility (e.g., confidential government works, proprietary datasets), authors have to ensure that their manuscript provides sufficient details to allow subsequent scholars to reimplement the proposed algorithm and attain the same results. This can be done with a combination of pseudocode and a detailed appendix. Key details to disclose include parameters,

layers, pre-processing steps, key operations, hardware setups, programming packages, and others. Zhu et al. (Forthcoming) provide an excellent example of how these details can be presented in a series of carefully designed appendices. While not the centerpiece of application-driven DL research, these details are often essential to facilitating scientific reproducibility.

*Guideline 6: Articulate clearly the deep learning contribution via the KCF for DL-ISR.* An essential component to executing academic research is clearly stating contributions. Within DL for ISR, this can manifest itself by positioning the contribution in the KCF. Contributions should be stated in a fashion that helps the audience understand and recognize the core novelty of the work (Gupta 2018). The provided details should also indicate how future work can build upon the proposed work to attain deeper levels of technical and domain novelty.

## Study

*Guideline 7: Reflect and summarize the key takeaways of the DL process.* A common critique of DL research in CS is the completion of the work following the technical experiments and contribution articulation (Gupta 2018). However, carefully studying the value and key takeaways of the proposed DL processes (e.g., architectures, encodings, learning paradigms, etc.) is essential for realizing the full potential of a DL-ISR study. In particular, authors should carefully reflect on their work and discuss how their work can apply to other disciplines and domains. These include generalizable or abstracted design principles or engineering considerations of their proposed research. Such details can help to improve the overall intellectual merit and novelty of the proposed work. Relatedly, authors should articulate the boundary conditions of their work. This can significantly assist in scope creep or incorrect applications of the proposed work.

*Guideline 8: Discuss any potential unexpected or unintended consequences of the proposed DL model and its implementation.* As indicated in the introduction, DL algorithms are often embedded

into systems that are deployed "at-scale." The large volume of users interacting with these systems increases the likelihood that the algorithm operates in a fashion unintended by the original developer. Recent examples include GAN-generated fake news, mass surveillance due to enhanced facial recognition, implicit racism in algorithm outputs, and many others. These issues can lead to significant loss of trust in AI-based systems, breach the privacy paradox, and ultimately create a digital divide in society. In light of these potentially severe societal ramifications, IS scholars should proactively indicate possible unexpected or unintended consequences or behaviors of their proposed DL processes. Two perspectives can be taken. First, researchers can look directly at instances where their proposed DL processes behaved in an unexpected fashion in their use case. Second, researchers can imagine how their proposed DL process would behave if it was embedded into a system deployed at scale. Taking both perspectives can help proactively identify the potential ramifications and boundary conditions of their proposed work. Moreover, they can serve as a strong basis for executing SBE-driven research (e.g., randomized field experiments) centered around tackling emerging themes of AI trust, AI privacy, and AI security.

*Guideline 9: State potential ethical concerns in how the DL approach was developed and in its potential usage.* Ethics are increasingly playing a larger role in academia. Within DL, there is significant focus on fairness, bias, transparency, and interpretability. Each area can significantly affect how a DL process is adopted, used, and governed in practice. To this end, IS scholars should carefully consider the potential ethical issues and impacts of their work. Key questions to consider include how the data were collected, biases implicit in the dataset, post-hoc analysis of the converged DL models to enhance interpretability and how the learning process behaves, and any ethical concerns that might come up when the model is used in practice and/or deployed at scale. Each detail can help facilitate the design of just DL systems, governance strategies, incentivizing usage, and motivate novel technical DL representations and learning strategies. These details can also drive grounded SBE-related research pertaining to how DL-based approaches are adopted and used across various conditions (e.g., technological, social, cultural, etc.).

**Synthesis**

*Guideline 10: Articulate how the proposed deep learning work can be translated into practice or industry.* There is significant focus across the academic landscape on attaining external grant funding, the key sources of which include major scientific funding agencies (e.g., NSF and NIH) and industry. However, attaining these sources of funding require clear articulations of the practical utility and value (i.e., broader impacts) of the work. Therefore, being able to clearly articulate the value of their work is essential for broader visibility at the federal levels. In the context of DL, a deeper discussion of how the environment and model outcomes together influence business and societal outcomes (as suggested in Figure 1) can help in the articulation of the translational process.

## LEVERAGING THE KCF TO IDENTIFY DEEP LEARNING RESEARCH OPPORTUNITIES FOR IS SCHOLARS

Identifying how IS scholars can make DL contributions requires an understanding of the current state of the field. To this end, we reviewed all deep learning papers published in prevailing IS journals and ICIS over the past five years (2016-2020). This time period represents DL's rapid proliferation across the academic and practitioner landscape. To execute this review, we first identified the top journals (including those from the "IS Senior Scholars Basket of Eight") and conferences with proceedings within the IS discipline. In alphabetical order, the journals selected were: *ACM Transactions on Management Information Systems* (TMIS), *Communications of the Associations of Information Systems* (CAIS), *Decision Support Systems* (DSS), *European Journal on Information Systems* (EJIS), *Information Systems Journal* (ISJ), *Information Systems Research* (ISR), *INFORMS Journal of Computing* (JoC), *Journal of Information Technology* (JIT), *Journal of Management Information Systems* (JMIS), *Journal of Strategic Information Systems* (JSIS), *Management Information Systems Quarterly* (MISQ), *Management Information Systems Quarterly Executive* (MISQE), and *Management Science* (MS). In addition to these prevailing journal venues, we also reviewed the top IS conference, the International Conference on Information Systems (ICIS).

For each venue, we searched for the term "deep learning" and its variations (e.g., neural networks, artificial intelligence) in the title, abstract, and keywords. Searching in this fashion retrieved articles where deep learning was the focus of the work. We then constructed a taxonomy that summarized the year, authors, venue, data, data types, tasks, DL models employed, learning paradigm, and their KCF contribution type. While the full details of this review are presented in Appendix A, we present a summary of the number and publication trends in Figure 2.

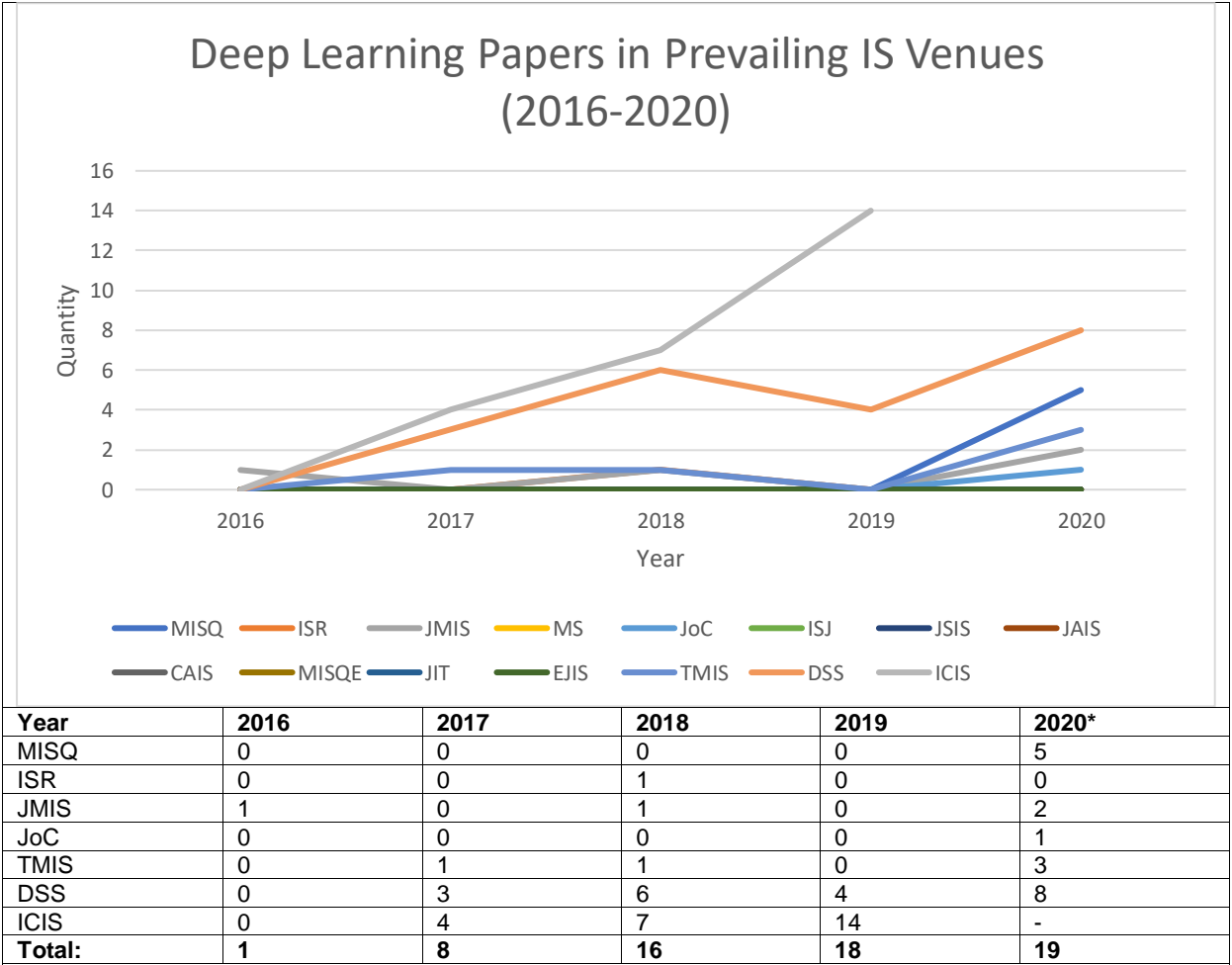

| Year | 2016 | 2017 | 2018 | 2019 | 2020* |
|------|------|------|------|------|-------|
| MISQ | 0 | 0 | 0 | 0 | 5 |
| ISR | 0 | 0 | 1 | 0 | 0 |
| JMIS | 1 | 0 | 1 | 0 | 2 |
| JoC | 0 | 0 | 0 | 0 | 1 |
| TMIS | 0 | 1 | 1 | 0 | 3 |
| DSS | 0 | 3 | 6 | 4 | 8 |
| ICIS | 0 | 4 | 7 | 14 | - |
| **Total:** | **1** | **8** | **16** | **18** | **19** |

**Figure 2. Number of Deep Learning Papers Published in Major IS Venues***

**\*Note:** At the time of this writing, ICIS 2020 papers had not been published.

In total, IS scholars have published 62 papers related to DL in the aforementioned venues. The venues with the highest number of DL-based papers are ICIS at 25, DSS at 21, ACM TMIS at five, MISQ at five, and JMIS at four. INFORMS journals within the University of Texas, Dallas (UTD) 24 list have lower numbers, with *Journal on Computing* having two, ISR one, and none at MS. The remaining journals (i.e., ISJ, JAIS, CAIS, JSIS, MISQE, JIT, EJIS) have no published or forthcoming DL papers. Conducting a similar search on the ACM Digital Library for the same five-year time period returned 86K results. The substantial difference between the number of papers produced by the IS community and related disciplines indicates that IS scholars have not fully leveraged the significant opportunity. This should not be an opportunity lost; IS scholars should quickly and proactively cultivate DL-ISR.

When growing the DL-ISR community, IS scholars should not directly aim to compete with CS. CS researchers often work in large labs that are deeply ingrained with technology giants such as Google, Facebook, Amazon, Tesla, Twitter, Netflix, and Apple. These relationships provide them excellent access to large datasets, computing resources, and engineering teams to execute fundamental DL research with large-scale and far-reaching impact (partially due to highly-visible, longstanding conferences with fast review cycles). These include general purpose image analysis, large-scale text analytics, massive pre-trained models (e.g., GPT-3), self-driving cars, Auto-DL, and others. This long-standing CS ecosystem significantly limits IS scholars from directly competing with CS scholars in conventional DL topics publicized by popular media, CS venues, or significant practitioner outlets (e.g., Gartner).

Despite these challenges, the unique positioning of IS within a business school provides the discipline with three key advantages over other disciplines engaging in DL research. First, IS scholars often have deep relationships with corporate partners in a wide range of industries (Rai 2020) ranging from start-ups to Fortune 100 companies. These relationships provide IS scholars with a significantly broader view of varying application areas, data, tasks, and requirements than the view afforded CS scholars. Second, IS scholars have a proven track record in forging inter-disciplinary relationships (e.g., health, psychology, cybersecurity, etc.) (Nunamaker et al. 2017; Bardhan et al. 2020). These ties provide a breadth of perspectives to inspire cross-cutting DL research. Finally, IS has historically served as the technical interface for conventional business school disciplines. These include accounting, finance, marketing, operations management, management, economics, supply chain management, and cybersecurity. Each discipline has vertical applications such as healthcare, retail and e-commerce, energy, transportation, and others. The combination of business function and application areas provides IS scholars a unique advantage in translating technical innovations into business or consulting practices (e.g., marketing, governance, policies, etc.) at a large-scale.

It is with these unique competitive advantages in mind that IS scholars can pursue high-impact DL-ISR. In particular, IS scholars can take the KCF, business functions, and application areas perspectives into account when positioning DL-ISR. Contributions can be made to a particular application directly, to a business function directly, or by taking both into account simultaneously. The KCF can serve as a vehicle to effectively gauge the type of contribution being made. Consideration of these perspectives has two key benefits. First, it avoids fundamental DL research that would put IS scholars in direct competition with CS. Second, these perspectives can facilitate "big tent," inter-disciplinary, and intra-disciplinary collaborations. As a result, it can limit labels of design science, economics, and behavioral research. These labels have specific connotations to IS researchers that can often curtail cross-tradition research (Rai 2018). Conceptually, each perspective can represent an axis on a three-dimensional plot (presented in Figure 3) to represent the possible contribution space for DL-ISR. Each point in this space is a possible contribution.

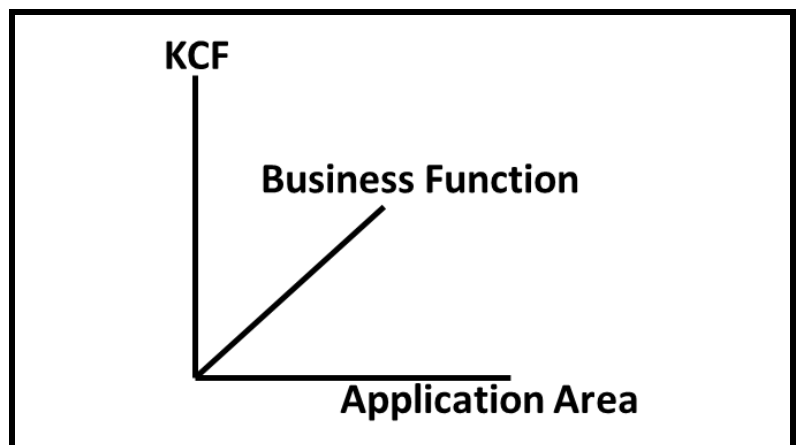

**Figure 3. Conceptual Space for DL-ISR Contributions**

Combining the KCF, business function, and application area perspectives in this fashion can lead to significant opportunities for IS scholars to make impactful DL contributions that are non-trivial. In practice, some regions of this space may have greater potential based on where the combination of data and problems make DL viable. To illustrate how these perspectives can be leveraged to formulate novel DL-ISR inquiries, we provide concrete exemplars in Table 5. The columns in the table combine application area and business functions, while the rows are the components of the KCF. Each cell is a potential contribution. We present four application areas and business function combinations: (1) healthcare in operations management, (2) retail and e-commerce in marketing, (3) government and policy in finance, and (4) software engineering in security. These business functions and applications are selected for several reasons. First, the applications listed have been stated by leading IS scholars as holding significant potential for major societal impact (Chen et al. 2012). Second, each business function has a tremendous amount of data that has yet to be tapped. Finally, each example requires both domain knowledge in the application as well as a deep understanding of how the business function is usually performed. Consequently, this can lead to novel DL research inspired by the unique characteristics of that application domain and/or business function.

**Table 5. Exemplars of KCF-Inspired Research Directions Across Applications and Business Functions**

| Contribution Type | Healthcare – Operations | Retail & E-Commerce – Marketing | Government and Policy – Finance | Software Engineering – Security |
|---|---|---|---|---|
| Domain-specific | Predicting near-term patient outcomes based on smartwatch and biometric monitoring | Predicting real-time couponing (i.e., product interest) for buyers based on in-store movements combined with data on prior purchase and online and in-store promotions | Predicting financial statement fraud for corporations | Detecting and predicting vulnerabilities in open source software with commercial vulnerability scanners and forecasting methods |
| Representation | Integrating effective representations involving audio (e.g., cough, speech etc.), longitudinal biometrics, social connections (e.g., relatives), and patient history from traditionally disparate data sources | Exploring effective representations of highly granular shopping cart routes with smart carts, sensor data from other in-store and online monitoring, and real-time data on customer responses to promotions shown | Exploiting representations of massive volumes of semi-structured information, inter-company networks, data on economic trends, and temporal performance of companies in similar industries to gauge outliers | Capturing relationships of applications and software across open-source coding repositories when detecting and predicting vulnerabilities |
| Learning | Designing exponential loss functions that excel in catching high-risk cases on time while minimizing false positives to save monitoring costs | Exploring online learning strategies that effectively capture consumer concept drift | Designing a novel, human-in-the-loop, active, and adversarial learning strategy since labeled fraudulent data availability is typically low and fraud often evolves to avoid detection | Incorporating adversarial robustness into the DL-based detection and prediction models to prevent adversarial poisoning attacks from nefarious inputs |
| System, Framework, or Workflow | Integrating DL-based patient monitoring into next-generation healthcare monitoring tools | Integrating DL-based couponing ideas into smart-carts in a retail environment | Integrating DL-based fraud monitoring engines into systems for auditing practice | Designing and deploying a novel system to enable security analysts to automatically identify and prioritize vulnerable software, key repositories, and users for advanced remediation |
| Innovation-Accelerating | Creating an anonymized open-source repository of patient-level longitudinal data and/or code from existing models | Making in-store movement data available in an anonymized format and code | Creation of artificial repositories of anonymized financial statements labeled based on intentional perturbation of key reporting metrics, along with DL code. | Providing open source vulnerability detection and prediction software |

The exemplars provided in Table 5 are threaded (i.e., build upon each other) for ease of exposition and do not suggest that a single paper should make such contributions across KCF dimensions. Rather, we envision significant differences in terms of how scholars approach opportunities in the business functions and applications space, and these differences will likely determine which cell(s) the contributions are in. More generally, Table 5 illustrates how the KCF can facilitate the systematic development of DL-ISR centered around high-impact applications for key business functions. Taking the Government and Policy application area and the Finance business function as an example, the KCF can provide the first roadmap for IS scholars to systematically develop and position high-impact DL-ISR for predicting financial statement fraud. Contributions can range from applying emerging algorithms to enhance detection, to more advanced representations of multi-modal data, to human-in-the-loop systems. Each contribution can draw upon emerging DL approaches as well as integrate methodologies that have long been at the heart of the IS discipline. Prevailing methodologies include building causal theories, assessing the economic impact of systems, user interface design, user acceptance, user adoptions, and others. Positioning DL-ISR in this manner incorporates multiple perspectives, paradigms (behavioral, design science, and econometrics), and methodologies synergistically without favoring any approach. As a result, the positioning of DL-ISR can have far-reaching implications. These include incentive capabilities, auditing, governance strategies, data sharing, mechanism design, and others.

## CONCLUSION

Deep learning has rapidly emerged as an essential component to modern AI. IS scholars have significant potential in making timely and high-impact contributions to this rapidly emerging topic. However, DL research within IS is still in its nascency. Clear, prescriptive guidance on how IS scholars can approach, examine, and make DL-oriented contributions is essential for ensuring the IS discipline stakes its rightful place as a major player in the DL landscape. To this end, this work aimed to make five sets of contributions. First, we systematically summarized the major components of DL in a novel DL-ISR schematic. Compared to the overview of DL commonly seen in CS textbooks (e.g., Goodfellow et al. 2016), the review of DL was instantiated within the context of IS, where technical DL processes are driven by key factors from an application environment. Second, we presented a novel KCF to provide IS scholars a roadmap on how to position their DL contributions effectively and clearly. The KCF holds significant promise in quickly accelerating the rate, depth, and breadth of DL-ISR. Third, we provided a set of guidelines to help IS scholars generate rigorous and relevant DL-ISR in a systematic, high-quality fashion. Fourth, we reviewed prevailing journal and conference venues to examine how IS scholars have leveraged DL for various research inquiries. As part of this review, we positioned their contributions vis-à-vis the proposed KCF. Finally, we provided a unique perspective on how IS scholars can formulate DL-ISR inquiries by considering the interplay of business function(s), application area(s), and the KCF. This perspective intentionally emphasizes inter-disciplinary, intra-disciplinary, and cross-IS tradition perspectives. Consequently, it can result in high-impact research that is significantly different from related disciplines (e.g., CS).

As with any seminal work, we intentionally bounded our coverage in several ways. First, we did not describe the mathematical underpinnings (e.g., formulations) of deep learning. Such details have been extensively reviewed in numerous textbooks and review papers. Reviewing them would have made our work tutorial in nature, and less specific to IS scholars. Second, our literature review did not include other IS conferences such as Americas Conference on Information Systems (AMCIS), European Conference on Information Systems (ECIS), Pacific Asia Conference on Information Systems (PACIS), or Hawaiian International Conference on Systems Sciences (HICSS). However, even if these venues were considered, it would be unlikely that they would have resulted in a volume comparable to the quantity present in the larger CS academic landscape. Finally, we opted to provide exemplars of future DL-ISR research directions, rather than providing a comprehensive list. This helps to avoid speculation or accidental research misdirection. Despite these bounds, we firmly believe that the DL-ISR schematic, KCF, and guidelines can provide IS scholars an unprecedented ability to advance the scale, scope, and impact of deep learning research to create a positive societal impact.

# APPENDIX A: SUMMARY OF DEEP LEARNING PAPERS PUBLISHED IN PREVAILING INFORMATION SYSTEMS VENUES

We reviewed all deep learning papers published in prevailing IS journals over the past five years (2016-2020). This time period represents DL's rapid proliferation across the academic and practitioner landscape. To execute this review, we first identified the top journals (including those from the "IS Senior Scholars Basket of Eight") and conferences with proceedings within the IS discipline. In alphabetical order, the journals selected were: *ACM Transactions on Management Information Systems* (TMIS), *Communications of the Associations of Information Systems* (CAIS), *Decision Support Systems* (DSS), *European Journal on Information Systems* (EJIS), *Information Systems Journal* (ISJ), *Information Systems Research* (ISR), *INFORMS Journal of Computing* (JoC), *Journal of Information Technology* (JIT), *Journal of Management Information Systems* (JMIS), *Journal of Strategic Information Systems* (JSIS), *Management Information Systems Quarterly* (MISQ), *Management Information Systems Quarterly Executive* (MISQE), and *Management Science* (MS). In addition to these prevailing journal venues, we also reviewed the top IS conference, International Conference on Information Systems (ICIS).

For each venue, we searched for the term "deep learning" and its variations (e.g., neural networks, artificial intelligence, etc.) in the title, abstract, and keywords. Searching in this fashion retrieved articles where deep learning was the focus of the work. After filtering out work that did not contain DL-related content, we constructed a taxonomy that summarized the year, authors, venue, data, data types, tasks, DL models employed, learning paradigm, and their KCF contribution type. While the main text presented the overall quantities and trends of papers across the venues, we present a complete summary of the contructed taxonomy in Table A1. Papers are first organized in alphabetical order based on their venue name, then based on reverse chronological order.

**Table A1. Summary of Published Deep Learning Research in Prevailing IS Outlets**

| Authors | Year | Venue | Data | Data Type(s) | Task(s) | Model* | Learning Paradigm | KCF Positioning |
|---|---|---|---|---|---|---|---|---|
| Zinoyeva et al. | 2020 | DSS | Social media data | Text | Detect antisocial online behavior | Hierarchical attention network | Supervised; transfer | Domain; Representation |
| Punia et al. | 2020 | DSS | Historical sale data | Sequential | Estimate the demand | DNN | Supervised | System |
| Wang et al. | 2020 | DSS | Crowdfunding information (e.g., entrepreneur profile, founder-generated content) | Text | Detect text content emphasis | LSTM | Supervised | System |
| Zhang et al. | 2020 | DSS | Hosts and listings on Airbnb (i.e., self-description, superhost status, verified status, response time, response rate, price, listing type, cancellation policy) | Text; Structural data | Predict guests' trust | BiLSTM | Supervised | Domain; System |
| Kim et al. | 2020 | DSS | Social media; essay score data; LibraryThing reviews | Text | Analyze text (classification and regression) | CNN | Supervised | Representation |
| Chaudhuri and Bose | 2020 | DSS | Geo-tagged images from earthquake-hit regions | Image | Identify survivors in images of debris after earthquake | CNN: ResNet 50, Inception, AlexNet | Supervised; transfer | Domain |
| Zhang and Mahadevan | 2020 | DSS | Flight trajectories | Trajectory | Predict flight trajectories | DNN; LSTM; Bayesian networks | Supervised; Bayesian | System |
| Kraus and Feuer | 2019 | DSS | Turbofan engine degradation simulation | Sensor signal | Predict the remaining useful life of equipment | Structured-effect neural network | Supervised; Bayesian | Representation |
| Park and Song | 2019 | DSS | Business process log | Sequential data | Predict the future performances of a business process | CNN, LSTM, Long-term Recurrent Convolutional Networks | Supervised | Domain; Representation |
| Vo et al. | 2019 | DSS | Yahoo finance data (including stock prices, environmental, social and governance metrics) | Sequential data | Predict stock returns | Multivariate BiLSTM | Supervised; reinforcement; multi-source | System |
| Kim et al. | 2020 | DSS | Start up profiles and patent abstracts | Text | Recommend startups as technology co-op candidates | Doc2vec model | Supervised | Domain |
| Kraus and Feuerriegel | 2019 | DSS | Financial disclsure | Text | Analyze sentiment of financial news to forecast stock price movements | LSTM | Supervised; transfer | Domain |
| Guan et al. | 2019 | DSS | Amazon data | Text; Images | Provide personalized recommendations | Autoencoder; Convolutional autoencoder | Supervised; multi-task and multi-source | Representation |
| Kratzwald et al. | 2018 | DSS | Literary tales; election tweets; ISEAR; headlines; general tweets | Text | Recognize emotion to support decisions (affective computing) | BiLSTM | Supervised; transfer | Domain |
| Loureiro et al. | 2018 | DSS | Historical sale data | Sequential data | Forecast sales in fashion retail | DNN | Supervised | Domain |
| Mahmoudi et al. | 2018 | DSS | Social media data | Text | Classify investor sentiment | CNN; RNN (LSTM, GRU) | Supervised | Domain |
| Zhang et al. | 2018 | DSS | Airbnb data (rating, reviews, | Text; structural | Predict guests' | DNN | Supervised | System |

| | | | response, rates, description, photos) | data; images | perceived trust in Airbnb | | | |
|---|---|---|---|---|---|---|---|---|
| Smadi et al. | 2018 | DSS | Phishing email data | Text | Detect phishing emails | DNN | Supervised; reinforcement | Domain |
| Walczak and Velanovich | 2018 | DSS | Pancreatic cancer data | Structural data | Predict the 7-month survival of pancreatic cancer patients | DNN | Supervised | Domain |
| Wang and Xu | 2017 | DSS | Vehicle insurance claims | Text; structureal | Detect automobile insurance fraud | DNN | Supervised | System |
| Evermann et al. | 2017 | DSS | BPI Challenge dataset | Text | Predict the next event in a business process | LSTM | Supervised | Domain |
| Kraus and Feuerriegel | 2017 | DSS | Financial disclsure | Text | Analyze sentiment of financial news to forecast stock price movements | LSTM | Supervised; transfer | Domain; Innovation-Accelerating |
| Chai et al. | 2019 | ICIS | Customer reviews | Text | Learn topic for each word; detect fake reviews | LSTM, VAE, topic models | Supervised; unsupervised; Bayesian | Representation |
| Qiu et al. | 2019 | ICIS | Alzheimer's disease data (MRI images, genetic markers and clinical notes) | Image; text; genetic data | Distinguish the transitional stage of Alzheimer's | Deep autoencoder | Supervised; unsupervised; multi-modal | Representation (Architecture) |
| Davcheva | 2019 | ICIS | Online mental health forum data | Text | Predict mental health symptoms and conditions | BiLSTM | Supervised; multi-label | System |
| Xie et al. | 2019 | ICIS | Patient discussions about drug use and recovery in an online health forum | Text | Classify a sentence as treatment barrier or not | SImilarity Network-based DEep Learning (SINDEL) | Supervised; multi-view | Representation (Embedding) |
| Yang et al. | 2019 | ICIS | Social media postings | Text | Detect senior executives' personalities | CNN-LSTM (character level CNN, BiLSTM and attention) | Supervised | Representation (Embedding) |
| Gu and Leroy | 2019 | ICIS | Electronic health records (EHR) | Text | Label free-text to create new labeled data; classify autism spectrum disorder | Ensemble of BiLSTM | Semi-supervised; multi-label; ensemble | System |
| Xu et al. | 2019 | ICIS | Posts and replies on online cancer community | Text | Classify patient needs and provide social support | Ensemble of LSTM and CNN | Supervised; multi-label; ensemble | System; Representation |
| Ghanvatkar and Rajan | 2019 | ICIS | Electronic health records (EHR) | Sequential data | Predict mortality for patients at intensive care units | Multi time scale RNN | Supervised | Domain; Representation |
| Kulkarni et al. | 2019 | ICIS | Customer online reviews | Text | Measure consumer brand engagement | LSTM; CNN | Transfer; supervised; | Domain; System |
| Li et al. | 2019 | ICIS | Service deal data from a service e-tailing website (including portal images, price, discount, etc.); Open Images dataset; Kinetics dataset | Image; structural data | Extract these experience-related stimuli from the portal images | Deep residual networks; faster R-CNN | Transfer; supervised | Domain; System |
| Guan et al. | 2019 | ICIS | Amazon data (including product description and reviews) | Text; image | Detect and recognize faces | Multi-task cascaded | Supervised | System |

| | | | | | | convolutional networks | | |
|---|---|---|---|---|---|---|---|---|
| Lysyakov et al. | 2019 | ICIS | Twitter data | Text | Classify tweets | LSTM | Supervised; transfer | System |
| Tofangchi et al. | 2019 | ICIS | Hotel booking dataset | Structural data; sequential data | Profile users for personalized recommendation | LSTM | Supervised; transfer | Domain; System |
| Deng et al. | 2018 | ICIS | Microblogs | Text | Classify sentiment | Deep neural network | Supervised; transfer | Representation |
| Hou et al. | 2018 | ICIS | Title images | Image | Extract useful features | DNN | Supervised | System |
| Xie and Zhang | 2018 | ICIS | Hospitalization records | Sequential data | Predict patient's 30-day readmission risk | Trajectory-aware LSTM | Supervised | Representation |
| Park and Kim | 2018 | ICIS | Crowdfunding project data | Text; image | Fix racial discrimination in online crowdfunding | CNN; GRU | Supervised; adversarial; transfer | Domain; Representation; System |
| Ahangama and Poo | 2018 | ICIS | User activity data (Amazon review dataset; MovieLens 1M dataset) | Structural data | Resolve user entities | Variational autoencoder | Bayesian; unsupervised | Domain; Representation; Innovation Accelerating |
| Qiao and Huang | 2018 | ICIS | Financial time series | Sequential data | Predict the focal variable of interests | Hierarchical stacking of LSTM | Supervised; ensemble | Domain; System |
| Utomo et al. | 2018 | ICIS | Electronic health records (EHR) | Sequential data | Recommend treatment(s) for patients in intensive care units | Value iteration network | Reinforcement | Domain; System |
| Qiu et al. | 2017 | ICIS | Food images | Image | Predict the food categories contained in the image | CNN | Supervised; multi-label | System |
| Guo et al. | 2017 | ICIS | Phone call data | Audio | Analyze phonetic and acoustic features | CNN; LSTM | Supervised; ensemble | System; Domain |
| Urbanke et al. | 2017 | ICIS | Online retailer transaction data | Transaction data | Predict whether the product will be returned | DNN | Supervised | Representation |
| Xie et al. | 2017 | ICIS | Online health platform data | Text | Capture medication nonadherence reason expression | Sentiment-enriched BiLSTM | Supervised | Representation; System |
| Zhu et al. | Forthcoming | MISQ | Sensor signals for activities of daily living | Sensor signals | Identify activities of daily living | CNN, LSTM | Supervised | System; Representation; Innovation-Accelerating |
| Shin et al. | Forthcoming | MISQ | Social media data | Text; image | Predict post's popularity | CNN; Feedforward neural networks | Supervised; transfer | System; Representation (Embedding) |
| Zhang and Ram | 2020 | MISQ | Social media data; environmental sensors; socioeconomic census; outpatient illness surveillance data | Text; image; structural data | Classify profile images | CNN | Supervised; transfer | System |
| Liu et al. | 2020 | MISQ | YouTube descriptions | Text | Extract medical terms | BiLSTM | Supervised | System |
| Li et al. | 2020 | MISQ | Behavioral research papers | Text | Search constructs in behavioral literature through creation of | Character-level CNN; BiLSTM | Supervised | System; Representation (Architecture, |

| | | | | | TheoryOn system | | | Embedding) |
|---|---|---|---|---|---|---|---|---|
| Adamopoulos | 2018 | ISR | Social media data | Text; graph | Classify sentiments; compare users based on their latent representations | Commercial DL sentiment analysis tool; Deepwalk | Supervised | System |
| Ebrahimi et al. | 2020 | JMIS | DarkNet markets | Text | Identify DarkNet market threats | SVM, LSTM | Supervised | System |
| Zhu et al. | 2020 | JMIS | Sensor signals for activities of daily living | Sensor signals | Identify activities of daily living | CNN | Supervised; transfer | System; Representation |
| Guo et al. | 2018 | JMIS | User check-in data | Check-in data | Learn better features for user preferences and for point-of-interest recommendations | Semi-RBM; RBM; Deep autoencoder | Unsupervised; transfer | Domain; Representation |
| Li et al. | 2016 | JMIS | Customer reviews | Text | Analyse sentiment | RNN | Supervised | System |
| Zhou et al. | 2020 | JoC | Social media data | Graph | Learn account embeddings | Deepwalk | Semi-supervised | System |
| Tao et al. | 2020 | TMIS | Customer reviews | Text | Extract aspect terms | A WordNet-guided continuous-space language model | Semi-supervised; transfer | Representation (Embedding); System |
| Unger et al. | 2020 | TMIS | CARS (user feedback, points of interest); Frappe dataset; Yelp | Structureal transaction data | Provide personalized recommendations | Deep context-aware recommendation models | Supervised | Representation |
| Wang et al. | 2020 | TMIS | Online health community posts | Text | Predict user posting activities in online health communities | Parallel LSTM | Supervised | System |
| Zhu et al. | 2018 | TMIS | Job application records | Text | Provide job recommendations | Person-job fit CNN | Supervised | Representation |
| Chen et al. | 2017 | TMIS | Stock market data (e.g., open, high, low and close prices and volume) | Sequential data | Forecast market movement under high-frequency scenario | Double-layer neural network | Supervised; ensemble | Representation (Architecture) |

**Note:** BiLSTM = Bidirectional Long Short-Term Memory; CNN = Convolutional Neural Networks; DNN = Deep Neural Network; GRU = Gated Recurrent Unit; LSTM = Long Short-Term Memory; R-CNN = Region-based Convolutional Neural Network; RBM = Restricted Boltzman Machine; RNN = Recurrent Neural Network; VAE = Variational Autoencoder

The total number of papers (62) is far lower than the quantity in the ACM Digital Library pertaining to deep learning (86K results). Consequently, there is significant room for growth and expansion of DL-ISR. We discuss the key findings from our literature search in the remainder of the appendix. We summarize the findings into three sub-sections: (1) Data, Data Type, and Task, (2) Model and Learning Paradigm Selection, and (3) KCF Positioning. In each sub-section, we briefly summarize the current state of the field, then highlight several promising directions for IS scholars.

### Data, Data Type, and Task

*Current State of the Field:* To date, IS scholars have primarily leveraged publicly accessible data that are commonly available to CS scholars. These include e-commerce data (Guan et al. 2019; Ahangama and Poo 2018; Kulkarni et al. 2019), social media data (Liu et al. 2020; Ebrahimi et al. 2020; Shin et al. Forthcoming), and varying types of healthcare data (Gu and Leroy 2019; Zhu et al. Forthcoming; Zhu et al. 2020; Davcheva 2019; Ghanvatkar and Rajan 2019). The prevailing data types being studied include text and images. As a result, most encoding structures have been primarily sequences or grids. While the majority of these data are often also easily available to CS researchers, selected IS researchers have formulated more business-oriented tasks to execute. These include monitoring customer reviews, analyzing sentiment, modelling financial time series, and other tasks.

*Future Directions for IS Scholars:* Staying within the realm of selecting datasets that CS scholars can easily access can pose significant risks for the IS discipline's viability as a major player in the deep learning landscape. Therefore, there is a critical need for IS scholars to closely examine emerging application areas and business functions that IS scholars have a unique competitive advantage in. Selected examples include high-impact topics in cybersecurity (e.g., Dark Web analytics, disinformation, automated defenses), privacy (e.g., privacy policy analysis, personally identifiable information analytics), and healthcare (e.g., fall risk assessment). In each of these areas, these is significant room for IS scholars to leverage inter-disciplinary perspectives to design new approaches for representing and encoding data.

## Model and Learning Paradigm Selection

*Current State of the Field:* Similar to the current Data, Data Type, and Task, most studies have leveraged conventional deep learning algorithms such as CNN, ANN, RNN, and LSTM to accomplish their selected task. Moreover, many approaches have employed the conventional supervised and unsupervised approaches for their tasks. Numerous studies have also employed transfer learning, primarily for leveraging massive pre-trained models from Google to train their proposed deep learning processes. Finer grained learning paradigms such as active, federated, adversarial, and multi-task, have limited to no presence in extant IS literature.

*Future Directions for IS Scholars:* Emerging approaches for representing, processing, and learning hold significant promise in executing increasingly complex problems within emerging application areas. For example, multi-source, multi-task, and adversarial learning can help process multiple datasets simulatanously to achieve advanced data fusion for cybersecurity or privacy analytics applications. Attention-based deep learning models also hold significant promise in opening the black-box of deep learning to enhance model interpretability. Increased model interpretability can guide iterative deep learning refinements and also support user evaluation and user adoption tasks. Additionally, IS scholars can explore the interplay of conventional machine learning models with emerging architectures and learning strategies (e.g., few-shot learning, self-supervised learning) to significantly enhance the performance of conventional machine learning models alone.

## KCF Positioning

*Current State of the Field:* Most contributions from IS scholars have been at the domain level or at the system level. This speaks to IS scholars' abilities to carefully understand the domain requirements of their deep learning process as well as articulate a series of inter-twined activities to solve a key problem within their domain. Selected recent research has aimed to improve the representation of their phenomena of interest to further advance their model outcomes (Zhu et al. 2020; Ebrahimi et al. 2020; Park and Song 2020). However, most research has leveraged deep learning procedures to improve the state of the art in the selected application area. Additionally, only three studies have made innovation-accelerating contributions (Zhu et al. Forthcoming; Ahangama and Poo 2018; Kraus and Feuerriegel 2017).

*Future Directions for IS Scholars:* Significant opportunities remain for IS scholars to attain representation and learning level novelties. To this end, IS scholars can leverage the unique data characteristics, application requirements, and/or kernel theories drawn from relevant social behavioral and economics (SBE) and computational literature to inspire their novel deep learning processes. In addition, future work can significantly expand the focus and emphasis on attaining innovation-accelerating contributions. These can include public code bases, release of datasets, and disclosure of key deep learning operational details (e.g., hyperparameters). Ultimately, these contributions can assist future researchers in rapidly expanding the speed, range, depth, and impact of their DL-ISR inquiries.